\documentclass[lettersize,journal]{IEEEtran}
\usepackage{amsmath,amsfonts}
\usepackage{algorithmic}
\usepackage{algorithm}
\usepackage{array}
\usepackage[caption=false,font=normalsize,labelfont=sf,textfont=sf]{subfig}
\usepackage{textcomp}
\usepackage{stfloats}
\usepackage{url}
\usepackage{verbatim}
\usepackage{graphicx}
\usepackage{cite}
\usepackage{amsmath,amsfonts}
\usepackage{bm}
\usepackage{multirow}
\usepackage{colortbl}
\usepackage{amssymb}
\usepackage{booktabs}
\usepackage{subcaption}
\definecolor{mygray}{gray}{.9}
\definecolor{realred}{RGB}{191,0,64}
\definecolor{realgreen}{RGB}{0,136,124}
\definecolor{realblue}{RGB}{0,124,239}
\usepackage{xcolor}
\usepackage{ulem}
\usepackage{color}
\newcommand{\cmark}{\ding{51}}%
\newcommand{\xmark}{\ding{55}}%
\usepackage{pifont}

\let\oldtwocolumn\twocolumn
\renewcommand\twocolumn[1][]{
    \oldtwocolumn[{#1}{
    \begin{center}
    \includegraphics[width=\textwidth]{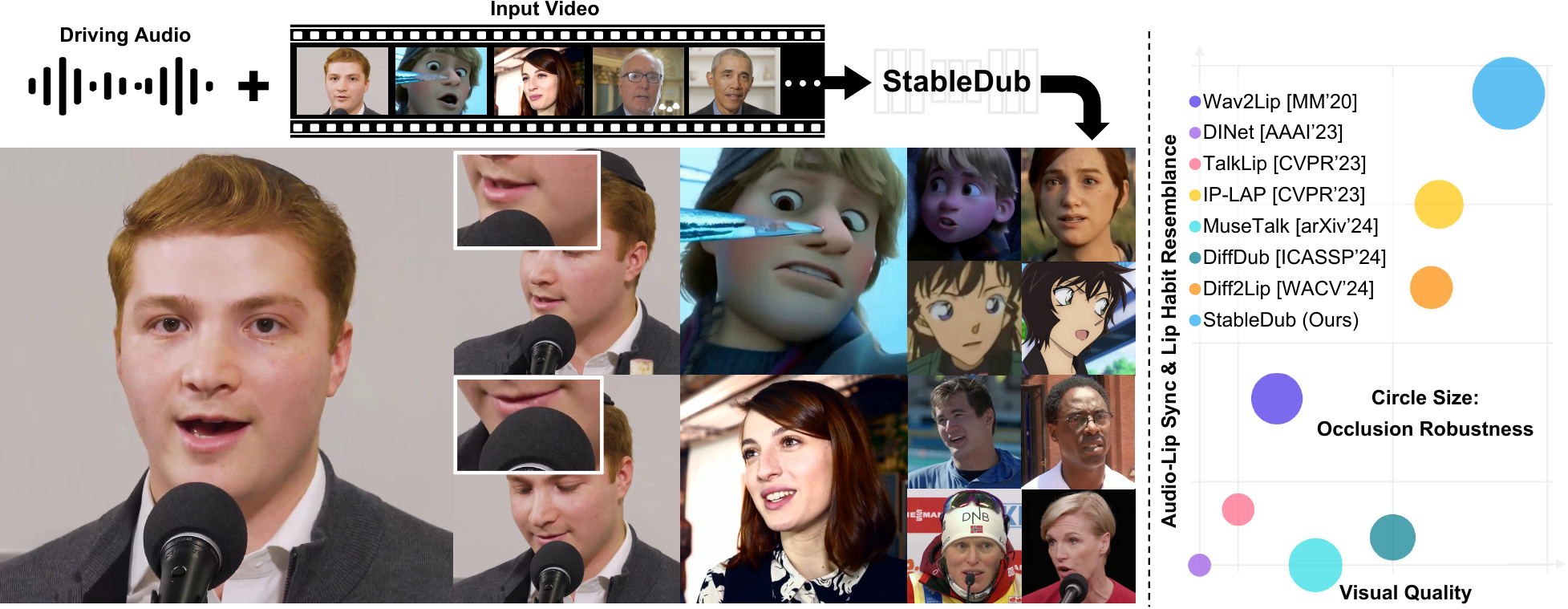}
    \captionof{figure}{We present StableDub for generalized visual dubbing (left side of the dashed line). Given a driving audio, StableDub alters the mouth region of the input video. The method can generate high-quality video frames and audio-synchronized lip movements that preserve the lip habits of the target avatars. It generalizes to human portraits, 2D/3D game or anime characters, even when their mouths are obscured by objects such as microphones. Extensive evaluation results (right side of the dashed line) demonstrate the superiority of StableDub compared with state-of-the-art visual dubbing methods.}
    \label{fig:teaser}
    \end{center}
    }]
}

\begin{document}

\title{StableDub: Taming Diffusion Prior for Generalized and \\ Efficient Visual Dubbing}

\author{Liyang Chen,
        Tianze Zhou,
        Xu He,
        Boshi Tang,
        Zhiyong Wu,~\IEEEmembership{Member,~IEEE,}\\
        Yang Huang,
        Yang Wu,
        Zhongqian Sun,
        Wei Yang,
        Helen Meng,~\IEEEmembership{Fellow,~IEEE}
  
\thanks{Manuscript under review.}}
%%%%%%%%%%%%%%%%

\maketitle

\begin{abstract}
The visual dubbing task aims to generate mouth movements synchronized with the driving audio, which has seen significant progress in recent years. However, two critical deficiencies hinder their wide application: (1) Audio-only driving paradigms inadequately capture speaker-specific lip habits, which fail to generate lip movements similar to the target avatar; (2) Conventional blind-inpainting approaches frequently produce visual artifacts when handling obstructions (e.g., microphones, hands), limiting practical deployment.
In this paper, we propose StableDub, a novel and concise framework integrating lip-habit-aware modeling with occlusion-robust synthesis. Specifically, 
building upon the Stable-Diffusion
backbone, we develop a lip-habit-modulated mechanism that jointly models phonemic audio-visual
synchronization and speaker-specific orofacial dynamics. To achieve plausible lip geometries and object appearances under occlusion, we introduce the occlusion-aware training strategy by explicitly exposing the occlusion objects to the inpainting process.
By incorporating the proposed designs, the model eliminates the necessity for cost-intensive priors in previous methods, thereby exhibiting superior training efficiency on the computationally intensive diffusion-based backbone. To further optimize training efficiency from the perspective of model architecture, we introduce a hybrid Mamba-Transformer architecture, which demonstrates the enhanced applicability in low-resource research scenarios.
Extensive experimental results demonstrate that StableDub achieves superior performance in lip habit resemblance and occlusion robustness. Our method also surpasses other methods in audio-lip sync, video quality, and resolution consistency. We expand the applicability of visual dubbing methods from comprehensive aspects, and demo videos can be found at https://stabledub.github.io.

\end{abstract}

\begin{IEEEkeywords}
Visual Dubbing, Talking Face Generation, Diffusion-based Video Synthesis
\end{IEEEkeywords}    
\section{Introduction}
% 介绍任务、阐述现状、三个关键点的解释和定义、现有方法的问题、难点（如下），所题方法
% 现在很多方法都是用audio-video的E2E路线，那么如何在这样的框架下，平衡三者的关系呢，还能高保证高效的训练效率呢

% audio-driven talking face generation 目的是生成面部动作与语音一致的视频，这项技术有着广泛的应用范围，如电影配音，游戏人物驱动、语音助手可视化，视频会议等。根据输入的形式和模型的泛化性能，该领域大致有三个方向，person-specific and person-agnostic visual dubbing 和 one-shot image animation。The visual dubbing 是对输入的一段视频的嘴部进行修改，person-specific setting 需要一定数量的视频对model进行finetune，而person-agnostic的method可以直接修改input video不需要finetune。one-shot image animation setting是仅仅输入一张图片，生成头部的所有运动，不仅是下半张脸。本文针对是person-agnostic visual dubbing。
Audio-driven talking face generation aims to produce videos where facial movements are aligned with the driving audio. This technology has a wide range of applications, such as movie dubbing, game character driving, and voice assistant visualization. Based on the form of input and the generalization performance of the model, this field can be categorized into three tasks: person-specific and person-agnostic visual dubbing, and one-shot facial image animation. The visual dubbing modifies the mouth of an input video clip. The person-specific setting \cite{lipsync3d_2021, synctalk_2024, adamesh_2025, sdnerf} requires a certain amount of video to finetune the model, while the person-agnostic setting \cite{wav2lip_2020,iplap_2023,Diff2Lip_2024, stablesync_2024, stylesync_2023, resyncer_2024, personaTalk_2024, AnyoneNet_2023} directly modifies the input video without finetuning. The one-shot facial image animation setting \cite{avct_2022,sadtalker_2023,dreamtalk2024,hallo_2024,vexpress_2024,emo_2024} generates the whole head motions from a single input image rather than only editing the lower-half face. This paper focuses on person-agnostic visual dubbing.

% despite the 现有的visual dubbing方法在generalization方面存在显著不足，具体表现为以下几个方面：首先，生成的视频视觉质量不稳定（instable visual quality），尤其是在处理高分辨率的视频输入时，现有方法难以保持一致的嘴部生成质量，导致视频输出的视觉效果不佳。其次，当人物嘴部被遮挡（如被话筒等物体遮挡）时，现有方法往往会生成明显的瑕疵，无法准确还原嘴部运动，影响了视频的自然性和连贯性。此外，对于多样化的portrait images（包括真人/游戏或动漫角色），现有方法生成的嘴形运动与其本人不像(resemblance)，无法满足不同风格和类型角色的需求。
\begin{figure}[t]
    \centering
    \includegraphics[width=0.45\textwidth]{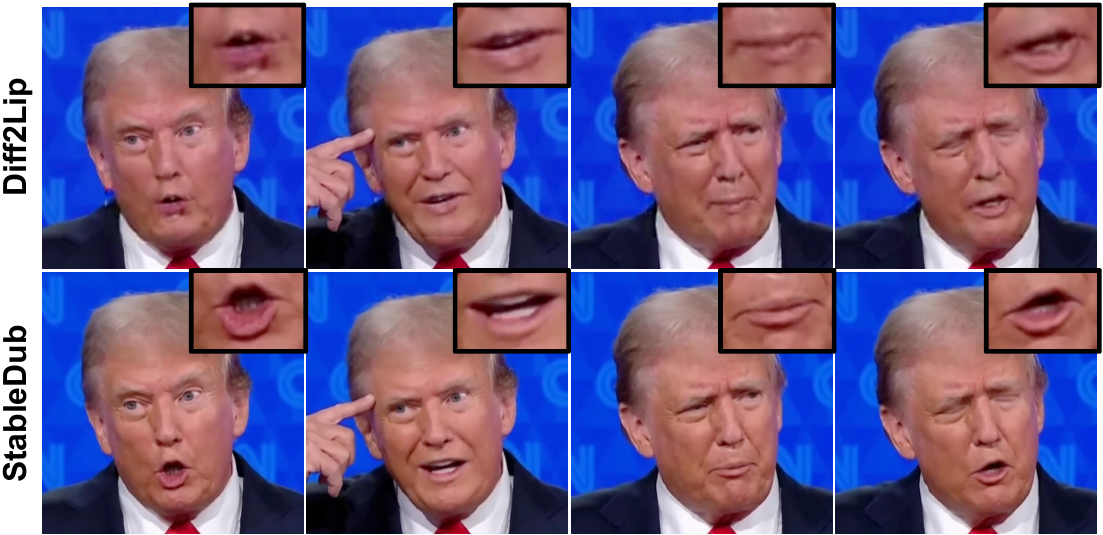}
    \caption{The previous method fails to generate lip movements that simulate the target avatar's habits, such as Donald Trump's distinctive lip habit, which is characterized by widely opening the mouth and puckering his lips.}
    \label{fig:trump}
\end{figure}

Despite the recent advantages in generating audio-synchronized lip movements, existing visual dubbing methods exhibit critical limitations in applicability, particularly concerning speaker-specific lip habit resemblance and occlusion robustness.
Firstly, existing methods fail to preserve idiosyncratic lip habits when animating diverse portrait images and speakers (including realistic human, game, or anime characters). As illustrated in Fig.~\ref{fig:trump}, Trump's lip habits often include a pronounced articulation, characterized by a widening mouth and puckering lips. Such distinctive articulatory signatures are inadequately captured by previous methods \cite{Diff2Lip_2024, diffdub_2024, styletalk2024}.  This limitation primarily stems from the synthesis pattern that exclusively rely on single-modal audio signals to drive lip movements, failing to incorporate speaker-dependent visual priors of facial dynamics and cross-modal articulation patterns inherent in individual talking behaviors.
Some recent works \cite{personaTalk_2024,styletalk2024,zhong2025,resyncer_2024} share a similar goal with ours by modeling person-specific speaking styles from a reference video. However, their reliance on strong facial priors \cite{iplap_2023, personaTalk_2024,stylesync_2023,zhong2025,realtalk_2024} or the need for fine-tuning on target speaker data \cite{resyncer_2024} limits their generalization and applicability.

Secondly, occlusion handling remains a persistent challenge for visual dubbing systems. 
Current methods show performance degradation when the mouth is partially or completely occluded by objects (e.g., microphones). It usually leads to implausible lip geometries and visual artifacts. 
This deficiency primarily arises from the
native blind inpainting paradigm that employs flawed facial masks to indiscriminately erase mouth regions, subsequently attempting reconstruction while disregarding the presence of occlusion objects.

To address these critical challenges, we propose \textbf{StableDub}, a novel and concise framework for generalized visual dubbing that integrates lip habit preservation with occlusion-robust synthesis. Building upon the U-Net-based Stable Diffusion backbone \cite{sd_2022}, we develop an audio conditioning and lip-habit-modulated mechanism that jointly models phonemic audio-visual synchronization and speaker-specific lip habits. Unlike prior audio-only methods \cite{diffdub_2024, difftalk_2023}, our method incorporates dual-modal audio and visual features for enhanced lip synchronization and lip habit resemblance. We introduce the occlusion-aware training strategy with augmented elaborate facial masks that amend the conventional blind inpainting. By strategically applying spatial patch infusion for the masked region during training, the model is explicitly exposed to occlusion objects and thus learns to reconstruct plausible lip geometries and object appearances. 

With the enhanced generalization capabilities enabled by these two designs, our method eliminates the necessity of cost-intensive priors in previous methods to ensure identity and lip-sync consistency, which significantly improves training efficiency for the computation-cost diffusion-based backbone. Furthermore, to optimize training efficiency in architecture design, we introduce a hybrid Mamba-Transformer architecture by integrating the emerging Mamba mechanism \cite{mamba_2024}, renowned for its linear-complexity state-space modeling, to replace the self-attention modules in spatiotemporal computation. This operation further enhances the applicability of visual dubbing methods in low-resource research scenarios.
Extensive experimental results demonstrate that StableDub outperforms the state-of-the-art (SOTA) methods in lip habit resemblance and occlusion robustness. StableDub also exhibits great generalizability in audio-lip sync, visual quality, and resolution consistency with the original video input.

% Talking face、SD
The main contributions can be summarized as follows:
\begin{itemize}
\item Lip habit-aware modeling: a dual-modal framework that synergizes audio-visual cues to preserve speaker-specific lip habits, overcoming the audio-only bias of prior methods by explicitly learning idiosyncratic orofacial patterns.
\item Occlusion-robust synthesis: a simple but effective object-aware inpainting process with the well-designed mask augmentation
that enables robust mouth reconstruction under occlusion scenarios.
\item Enhanced training efficiency: the obtained generalization capability permits removal of redundant priors, while Mamba-based complexity optimization achieves faster spatiotemporal convergence without quality degradation.
\item Enhanced applicability: Fig.~\ref{fig:teaser} and Tab.~\ref{tab:diff}, as well as extensive experiments, demonstrate that StableDub outperforms the state-of-the-art methods in various aspects and expands the applicability of visual dubbing methods.
\end{itemize}

\section{Related Works}
% \begin{table}[t]    
%     \centering
%     \resizebox{0.5\textwidth}{!}{
%         \begin{tabular}{lccccc}
%             \toprule
%             \multirow{2}{*}{Method} & \multicolumn{2}{c}{Generalization} & \multicolumn{3}{c}{Efficiency}\\
%             \cmidrule(lr){2-3} \cmidrule(lr){4-6}
%             & OR & LHR & ExtraPrior & TrainHours & Speed/Mem \\
%             \midrule
%             Diff2Lip &\xmark &\xmark &SyncExpert	&32/A6000	&0.78s/5.5GB\\
%             Hallo &\xmark &\xmark &ReferenceNet  &40/A100	&1.71s/10GB\\
%             Ours &\cmark &\cmark &- &16/V100	&0.62s/5GB\\
%             \bottomrule
%         \end{tabular}
%     }
%     % \vspace{-10pt}
%     \caption{Difference with state-of-the-art methods. Our method achieves superior generalization on lip habit resemblance (LHR) and occlusion robustness (OR) while demonstrates enhanced training efficiency without the dependency on extra priors of ReferenceNet \cite{hallo_2024} and SyncExpert \cite{wav2lip_2020}. The inference requires 5 GB of GPU memory and takes 0.62 seconds per frame when generating a 16-frame clip.}
%     % \vspace{-12pt}
%     \label{tab:diff}
% \end{table}
\begin{table}[t]    
    \centering
    \caption{Difference with state-of-the-art methods. Our method achieves superior generalization on lip habit resemblance (LHR) and occlusion robustness (OR) while demonstrates enhanced training efficiency without the dependency on extra priors of ReferenceNet \cite{hallo_2024} and SyncExpert \cite{wav2lip_2020}. The inference (Inf) requires 5 GB of GPU memory (Mem) and takes 0.62 seconds per frame when generating a 16-frame clip.}
    \resizebox{0.5\textwidth}{!}{
        \begin{tabular}{lcccccc}
            \toprule
            \multirow{2}{*}{Method} & \multicolumn{2}{c}{Generalization} & \multicolumn{4}{c}{Efficiency}\\
            \cmidrule(lr){2-3} \cmidrule(lr){4-7}
            & OR & LHR & ExtraPrior & TrainHours & Inf.Speed & Inf.Mem \\
            \midrule
            Diff2Lip &\xmark &\xmark &SyncExpert	&32/A6000	&0.78s &5.5GB\\
            Hallo &\xmark &\xmark &ReferenceNet  &40/A100	&1.71s &10GB\\
            Ours &\cmark &\cmark &- &16/V100	&0.62s &5GB\\
            \bottomrule
        \end{tabular}
    }
    \label{tab:diff}
\end{table}
\begin{figure*}[t]
    \centering
    \includegraphics[width=0.9\textwidth]{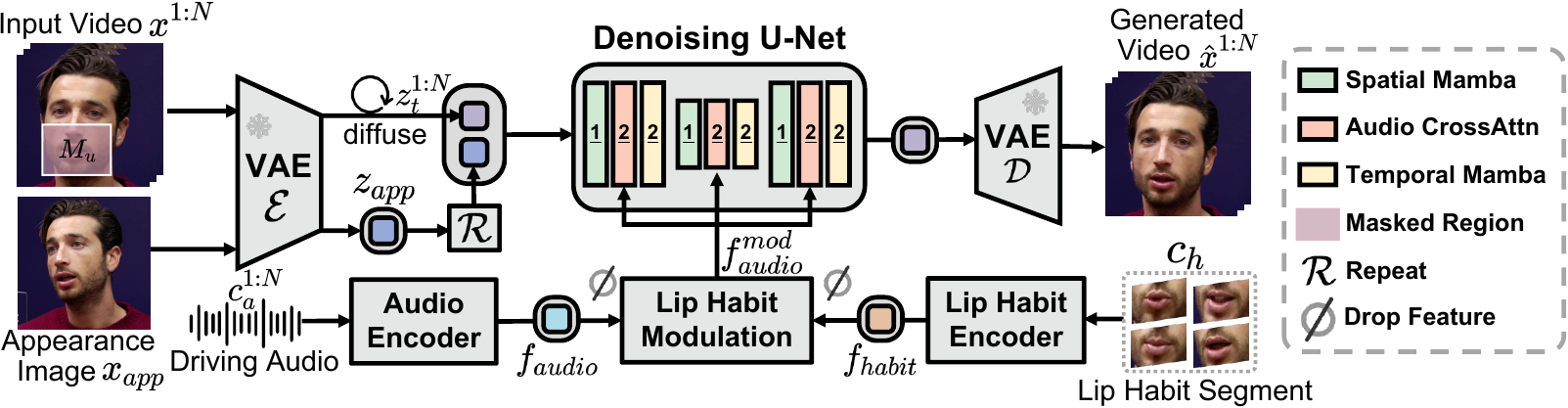}
    \caption{Given the input video of the target avatar and arbitrary driving audio, StableDub alters the masked mouth region and generates lip movements that match the audio content. To supplement appearance details in the generation process, we concatenate an appearance image with per-frame video latent and send them into the denoising U-Net. To enhance the resemblance of generated lip movements with the target avatar, we extract the habit feature from another lip segment and modulate it with the audio feature. The modulated feature is sent to the audio cross-attention layer in each U-Net block. The spatial and temporal Mamba layers are incorporated to efficiently ensure image quality and video coherence. \underline{1} or \underline{2} denotes that the layer is optimized in the training of stage 1 or stage 2.}
    \label{fig:fig2}
\end{figure*}
\noindent \textbf{Audio-Driven Talking Face Generation.}
% 最近有很多工作\cite{sadtalker, emo_2024, hallo_2024}投入到one-shot facial image animation上取得了令人瞩目的效果. 他们通过语音预测整个头部运动，主要应用在短视频制作娱乐，对真实性的要求不高。visual dubbing 则是只编辑一段视频的嘴巴部分，对真实性的要求更高，被用在更加严肃的场合，如视频翻译和虚拟人直播。以往的工作为了保证真实性，通常会在某个特定人物上训练或finetune，得到足以媲美真人的效果。随着模型能力的增强和数据来源更加的广泛，person-agnostic技术逐渐成熟。一些方法利用GAN来生成保证图像生成的质量，利用lip-sync expert学习audio-to-lip mapping。近期也有一些方法利用diffusion来进一步提高生成画面的质量。以上提到的这些方法在visual quality和lip sync做了很多尝试，但是忽略了对generalization的全面考虑。大部分方法定义generalization是模型naturally generalize across different identities，但是没有考虑实际应用中的人脸所处的场景，尤其是有物体遮挡嘴部时。本文相比于先前的工作，不光是在visual quality上有所提升，更对generlization的范围进行了扩展，致力于研究更加stable 的visual dubbing方法。
Many works \cite{sadtalker_2023,dreamtalk2024,emo_2024,vexpress_2024,hallo_2024} about one-shot facial image animation achieve remarkable results, and are mainly applied for entertainment. The requirement for realism is not stringent.
Visual dubbing, on the other hand, is applied in more serious scenarios such as video translation and virtual human live streaming. Previous works \cite{lipsync3d_2021} are usually trained or finetuned on a specific person to achieve realism comparable to real recordings. With the enhancement of model capabilities and the broader availability of data sources, person-agnostic technology has gradually matured. Some methods \cite{dinet_2023,talklip_2023, stylesync_2023, stablesync_2024, resyncer_2024} use generative adversarial nets (GANs) to improve image quality and employ lip-sync experts \cite{syncnet_2017, stablesync_2024} to learn audio-to-lip mapping. Recently, some methods \cite{difftalk_2023,diffdub_2024,Diff2Lip_2024, personaTalk_2024} introduce diffusion to further improve the quality of generated images. Some latest methods focus on personalized avatar generation by extracting person-specific styles from facial landmarks \cite{iplap_2023} or expression parameters\cite{personaTalk_2024,styletalk2024,zhong2025,resyncer_2024}, but these handcrafted features provide limited representations and cannot fully capture complex facial dynamics and texture variations. The aforementioned methods mostly focus on visual quality, lip sync, and speaking styles, but lack comprehensive consideration of generalization. Most methods \cite{stylesync_2023,videoretalking2022,difftalk_2023} define generalization as the model naturally generalizing across different identities, but they rarely consider diverse application scenarios, especially when objects occlude the mouth. Compared to previous works, this paper not only improves visual quality and audio-lip sync but also expands the scope of generalization.

\noindent \textbf{Diffusion-Based Video Generation.}
% 最近利用latent diffusion model 进行text-to-video任务获得了重大进展。以往的方法通常 train the image generative model in a latent space of reduced computational complexity，然后将temporal layers嵌入到预训练的结构中实现视频片段的生成。最近一些工作也将类似的结构adapt到one-shot image animation 上，获得了promising的结果。本文follow以前的工作也采取了这样的framework，但是将这个frmework与visual-dubbing任务的的需求相融合，获得了更高效的训练和定制化的结果。
Recently, significant progress has been made in text-to-video tasks using latent diffusion models. Previous methods \cite{animateanyone_2024,animatediff_2023} typically train the image generative model in a latent space of reduced computational complexity, and then embed temporal layers into the pre-trained structure to generate video clips.
Some works \cite{magicanimate_2024,emo_2024,hallo_2024,vexpress_2024} adapt similar structures for one-shot image animation, achieving promising results. This paper follows previous works and adapts to the visual dubbing task, achieving efficient training and customized results.

\noindent \textbf{Efficient Generation with Mamba.}
While diffusion models demonstrate remarkable capabilities, the training costs create accessibility barriers for many researchers. Mamba \cite{mamba_2024}, a novel state space model (SSM), garners widespread attention due to its powerful ability to model long sequence dependencies and high computational efficiency.
Mamba introduces a selection mechanism to structured SSM for context-dependent reasoning. Compared to Transformers \cite{transformer_2017} based on quadratic-complexity attention, Mamba excels at processing long sequences with linear complexity. We utilize Mamba to reduce the computational cost for the visual dubbing task.

\noindent \textbf{Remark.}
Our work diverges from previous methods by pursuing a more stable visual dubbing framework through expanded generalization scope and optimized training efficiency, thereby advancing the practical applicability of visual dubbing methods in diverse real-world scenarios.
\section{Method}
\label{sec:method}
In this section, we present StableDub to achieve generalized and efficient visual-dubbed video generation. 
As shown in Fig.~\ref{fig:fig2}, given the driving audio $c_{a}^{1:N}\in\mathbb{R}^{ N \times d}$ and the input video $x^{1:N}\in\mathbb{R}^{ N \times c \times h \times w}$, we randomly select a frame as the appearance image $x_{app}\in\mathbb{R}^{c \times h \times w}$, and sample another segment $c_{h}\in\mathbb{R}^{K \times c \times h \times w}$ and crop the lip region to provide the lip habit information.
StableDub aims to generate the audio-lip synced video $\hat{x}^{1:N}$ by inpainting the masked mouth region of the input video.
We first describe the preliminaries about the latent diffusion model (LDM) in Sec.~\ref{sec:ldm}, which is the foundation of the proposed method.
We demonstrate the exploration of lip habit modeling in Sec.~\ref{sec:habit}, the design of occlusion-aware inpainting and the training and inference strategy in Sec.~\ref{sec:loss}, and the network efficiency in  Sec.~\ref{sec:Mamba-Transformer}.

\subsection{Preliminaries}
\label{sec:ldm}
The latent diffusion model performs the diffusion process in the compact latent space rather than the original high-dimensional image space, which greatly reduces computational complexity while maintaining generation quality. Given an input RGB image $x_0$, a pretrained VAE encoder $\mathcal{E}$ maps it into latent feature $z_0 = \mathcal{E}(x_0)$. The forward diffusion process is then applied to produce $z_t$ by adding Gaussian noise: 
\begin{align}
z_t=\sqrt{\bar{\alpha_t}}z_0+\sqrt{1-\bar{\alpha_t}}\epsilon,\mathrm{~}\epsilon\thicksim\mathcal{N}(0,I),
\end{align}
where $t\in\{1,2,\ldots,T\}$ and $\bar{\alpha_t}$ denotes the noise strength at step $t$. A conditional trainable U-Net-based network $\epsilon_\theta$ then learns the denoising process by predicting the added noise. The training objective of the conditional LDM is defined as
\begin{align}
\mathcal{L}_{LDM}=\mathbb{E}_{z_0, c, \epsilon, t}[\|\epsilon-\epsilon_\theta(z_t, t, c)\|_2^2],
\end{align}
where $c$ is the condition. 
At inference, $z_T$ is sampled from random Gaussian distribution and is progressively denoised and restored to $\hat{z}_0$ via deterministic sampling process. $\hat{z}_0$ is then decoded by the VAE decoder $\mathcal{D}$ to reconstruct the image $\hat{x}_0 = \mathcal{D}(\hat{z}_0)$.
The proposed model extends a pre-trained conditional LDM \cite{sd_2022} for image generation to video generation by introducing temporal layers and audio-related layers. 
% As shown in Fig.~\ref{fig:fig2}, the appearance image and the input video are processed through the VAE encoder to obtain latent features and then repeated along the temporal dimension. The repeated $z_{app}$ is concatenated with the diffused latent of the input video along the channel dimension and then sent into the denoising U-Net. The audio-related feature

\subsection{Lip Habit Modulated Audio Feature}
\label{sec:habit}
\begin{figure}[t]
    \centering
    \includegraphics[width=0.45\textwidth]{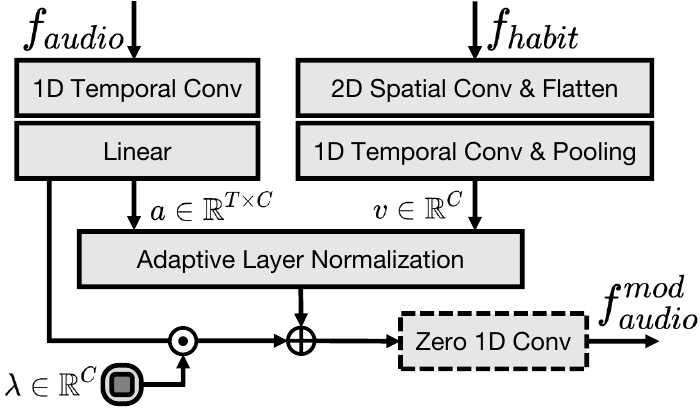}
    \caption{Lip habit modulation module transfers the lip habit to the audio feature. The output is connected to the denoising net with zero 1D convolution for better initialization.}
    \label{fig:adaLN}
\end{figure}
% 学习了残差信息，考虑了video的pattern，更好的承载lip habit
% 残差的结构辅助了lip habit的学习 （待定）
% adaLN 传统被利用在image style transfer上
% 不用habit呈现出over-smoothing、averaged
As shown in Fig.~\ref{fig:trump}, generating lip movements consistent with the target avatar's lip habit is crucial for realistic visual dubbing. To enhance lip habit resemblance and ensure synchronization, we propose a lip habit modulation module. We extract audio features $f_{audio}$ with Wav2Vec \cite{wav2vec_2019} audio encoder and image-level lip habit features $f_{habit}$ with VIT-based \cite{vit_2021} lip encoder. 
As shown in Fig.~\ref{fig:adaLN}, they are processed separately by convolution and pooling layers to obtain $a \in \mathbb{R}^{ T \times C}$ and $v \in \mathbb{R}^{C}$. Note that the input audio feature and lip habit segment are sampled from distinct segments in a video clip, and the lip-related feature $v$ is a global vector. Thus $v$ does not contain any audio-related information and is primarily optimized to represent lip movement habits.
The adaptive layer normalization (AdaLN) \cite{dit_2023} is then employed to modulate the audio signal in accordance with the lip habit, which can be formulated as
\begin{align}
\mathrm{AdaLN}(a,v)=(\frac{a-\mu(a)}{\sigma(a)})\mathrm{MLP_1}(v)+ \mathrm{MLP_2}(v),
\end{align}
where MLP denotes linear layer. This lightweight module is inspired by AdaLN for image style transfer \cite{styletransfer_2017}. 
To further facilitate the audio-synced generation, the AdaLN output is added with $a$ multiplied by a learnable vector $\lambda$. We also experimented with injecting lip habits via cross-attention, but found that AdaLN achieves comparable performance while requiring significantly lower memory and computational cost. This observation is consistent with findings from concurrent work \cite{magicmirror_2025} that utilize AdaLN for facial identity customization.

The output feature $f_{audio}^{mod}$ is injected into the denoising net by computing cross-attention with the latent denoising feature $z_t$. This computation can be formulated as
\begin{align}
\text{CrossAtten}(z_t, f_{audio}^{mod}) = softmax(QK^{\top} / \sqrt{d_k})V,
\end{align}
where $Q=W_{Q}z_t$, $K=W_{K}f_{audio}^{mod}$, $V=W_{V}f_{audio}^{mod}$, and $d_k$ denotes the latent dimension for cross attention. The zero-initlized 1D convlutional layer is for better initilization at the beginning of training without damaging the backbone performance. Here we adopt attention instead of Mamba since we empirically find that a hybrid Mamba-Attention network achieves better performance. Recent studies on the Mamba architecture \cite{mambavision_2024} have similar conclusions.
In the training stage, we randomly drop the audio and lip habit features with a drop probability of 0.1. This is to ensure robust audio-synced lip generation and lip habit learning. During inference, the lip habit segment is random sampled and shuffled from the input video and will not damage the lip sync. 

\subsection{Occlusion-Aware Inpainting}
\label{sec:loss}
\begin{figure}[t]
    \centering
    \includegraphics[width=0.45\textwidth]{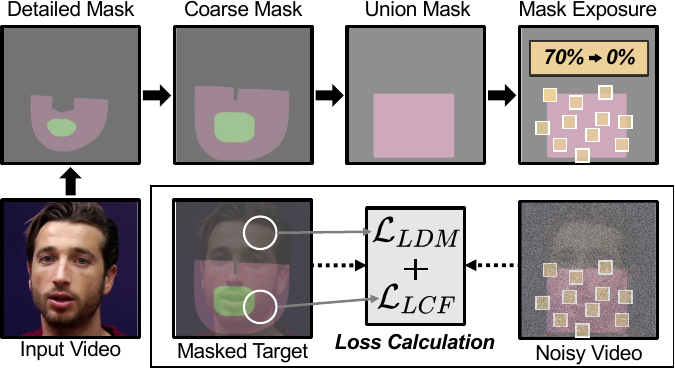}
    \caption{Illustration of facial masks. For the mask exposure operation, yellow patches denote unmasked regions exposed to inpainting, enabling occlusion-aware training. We define the ratio between the exposed patch area and the union mask area as the exposure rate. It is initialized at 70\% and progressively decreased to 0\% during training.}
    \label{fig:mask}
\end{figure}
\noindent \textbf{Facial Masks.}
As shown in Fig.~\ref{fig:mask}, we derive the coarse masks and union mask to facilitate the denoising process.
Most previous works \cite{wav2lip_2020,talklip_2023,musetalk_2024} independently detect the tight face bounding box in each frame of the video as the input region and mask the lower half of the box for inpainting.
Even when adopting temporal modules, this operation still damages the continuity of the generated video, especially in cases of static occlusions, because the input video frames have noticeable jumps.
StableDub first detects face landmarks for each frame to obtain a detailed mask, which precisely locates the lip and jaw. Then, we erode each detailed mask to obtain a coarse mask $M_c$ to eliminate the mouth estimation inaccuracy \cite{mediapipe_2019, stylesync_2023}. We union the height and width of each $M_c$ in the video segment and blurred the boundaries, resulting in a union facial mask $M_u$ that only indicates the approximate position of the mouth. This removes the jaw movement related to audio. Some works \cite{emo_2024, hallo_2024,stylesync_2023} take similar union operations.
The inputs to the denoising net are frames with fixed and much larger boundaries expanded by $M_u$, which ensures that the input frames are a continuous signal. When the mouth is occluded by an object, larger-area and multi-frame inputs can assist the network in predicting the shape of the object in the masked region from the surrounding areas.

\noindent \textbf{Progressive Mask Exposure.}
% 虽然扩大生成的区域可以让网络更多的感知object存在，但是inpainting的过程仍然是blind的，即mask区域的object长什么样子是不知道的。为此我们提出了一个简单的操作，在训练过程中把遮挡物主动暴露给inpainting过程。对于图五中的union mask，我们对其划分patch，并以随机取消一定比例的patches的mask。这些取消mask的区域有可能会暴露出遮挡物的部分样貌，使得inpainting 从这些patch猜测到遮挡物的样貌。我们使用progressive的mask暴露策略引导occlusion-robust lerning。
% 训练的一开始将暴露的比例设置为70%，网络通过copy-paste的方式减小学习难度；随着训练的进行，这一比例逐渐调整为0，网络从未被mask的其他区域猜测到object的appearance。
Although expanding the inpainted regions enables the network to better perceive object existence, the inpainting process remains inherently blind since the visual appearance of objects in masked regions remains unknown. To address this, we propose a simple yet effective operation that actively exposes occluders to the inpainting process during training. For the union mask shown in Fig. \ref{fig:mask}, we partition it into patches and randomly unmask a certain percentage of these patches. The unmasked regions may reveal partial appearances of occluders, enabling the inpainting process to infer their complete appearance from these visible clues. We implement a progressive mask exposure strategy to guide occlusion-robust learning. Initially setting the exposure ratio to 70\%, the network learns through copy-paste operations to reduce learning difficulty. As training progresses, this ratio gradually decreases to 0\%, forcing the network to deduce object appearances from other regions elsewhere in the image.

\noindent \textbf{Loss Function.}
Only the denoising loss $\mathcal{L}_{LDM}$ may not contribute to accurate lip movements \cite{diffusehead_2024,vexpress_2024}. 
Previous methods usually employ a lip-reading expert \cite{wav2lip_2020, talklip_2023, Diff2Lip_2024} or computing spatial loss in the full pixel space \cite{echomimic_2024, Diff2Lip_2024}. 
To avoid additional training overhead, we impose constraints in the latent space of the VAE, as we find that the VAE-encoded features preserve the spatial relationships of the original image. In the meantime, to enhance the attention of the denoising net on the lower-half face, we propose the latent coarse facial loss with the coarse facial mask $M_{c}$:
\begin{align}
\mathcal{L}_{LCF}=\mathbb{E}[\|&\epsilon-\epsilon_\theta(\bar{z}_t^{1: N}, t, c_{a}^{1:N},c_{h})\|_2^2\otimes M_c] \\ \notag
&\bar{z}_t^{1: N}=(z_t ^{1: N} \otimes M_u) \textcircled{c} z_{app},
\end{align}
where $\otimes$ is Hadamard production and $\textcircled{c}$ is frame-wise concatenation. $z_{app}$ denotes the VAE latent of the appearance image.
The total training loss can be written as:
\begin{align}
&\mathcal{L}_{total}=\mathcal{L}_{LDM} + \mathcal{L}_{LCF}.
\end{align}

\noindent \textbf{Classifier-Free Guidance.} We randomly drop audio and lip conditions in training. During inference, we introduce two guidance scales $\lambda_{a}$ and $\lambda_{h}$ and separately replace each condition with zero embedding $\varnothing$  to trade off the impact of these two control signals:
\begin{align}
\hat{\epsilon}_\theta(\bar{z}_t, c_a,c_h) &= \epsilon_\theta(\bar{z}_t, \varnothing,\varnothing)\\ \notag
&+\lambda_{a}(\epsilon_\theta(\bar{z}_t, c_a,\varnothing) - \epsilon_\theta(\bar{z}_t, \varnothing,\varnothing)) \\ \notag
&+\lambda_{h}(\epsilon_\theta(\bar{z}_t, c_a,c_h) - \epsilon_\theta(\bar{z}_t, c_a,\varnothing)).
\end{align}

\subsection{Efficient Network}
\label{sec:Mamba-Transformer}
% \begin{figure}[t]
%     \centering
%     \includegraphics[width=0.38\textwidth]{fig/mamba.pdf}
%     \caption{Mamba.}
%     \label{fig:mamba}
% \end{figure}
The introduced lip-habit modulation and occlusion-aware inpainting significantly enhance the generalization capability. In experiments, we empirically find that even when removing the prior modules for identity consistency and lip sync adopted in previous works, the model maintains robust preservation of facial identity across diverse input avatars while achieving precise audio-visual sync with various speech inputs. This architectural simplification substantially improves the training efficiency of the computationally intensive diffusion-based model without compromising generation quality.

\begin{figure}[t]
    \centering
    \includegraphics[width=0.49\textwidth]{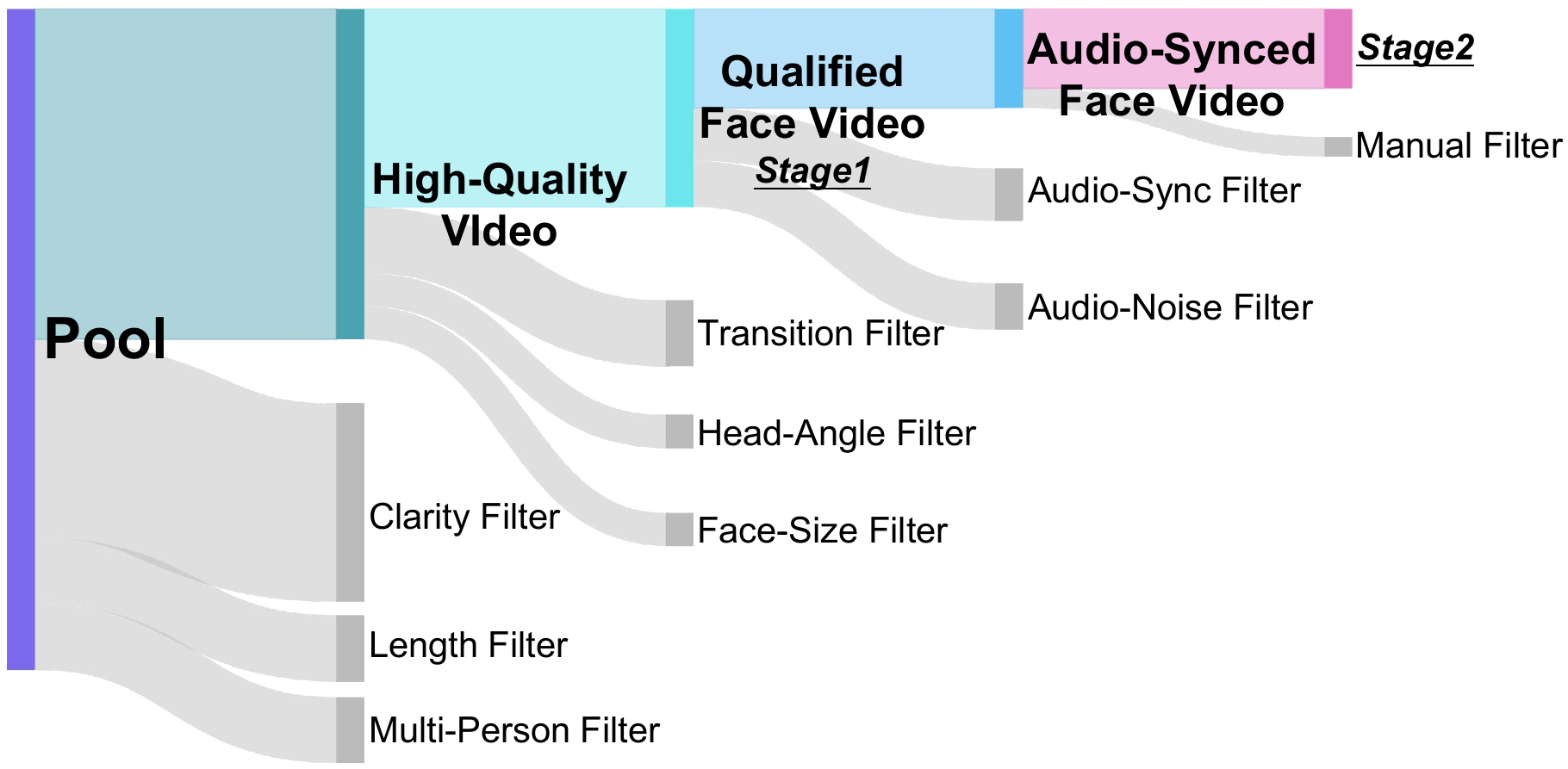}
    \caption{Illustration of data filtering and preprocessing. Qualified face videos without audio can be utilized in training of stage 1, while audio-synced videos are utilized in stage 2.}
    \label{fig:process}
\end{figure}
\noindent \textbf{Removal of ReferenceNet.} 
Recent SD-based methods \cite{emo_2024,hallo_2024} adopt a Dual U-Nets architecture. They try to extract the appearance by a cloned referenceNet, and inject it into the backbone U-Net for identity-preserved video generation.
However, it introduces additional unnecessary optimization parameters. This paper removes the referenceNet.
Besides, text input in the original Stable-Diffusion backbone is also removed, as this information is already provided by the appearance image. This removal reduces the network parameter count from 1.6B to 987M, significantly decreasing the parameter count by 40\%.

\noindent \textbf{Removal of SyncExpert.} For lip sync assurance, conventional visual dubbing methods rely heavily on SyncExpert \cite{wav2lip_2020}, as it significantly improves audio-lip sync on objective metrics. However,  SyncExpert introduces two critical limitations: (1) SyncExpert scores conflict with human perceptual judgments, and (2) SyncExpert's requirement to backpropagate losses at the RGB level complicates integration with latent-space-based diffusion models while incurring substantial training overhead. 
This work eliminates SyncExpert dependencies by directly computing loss in the latent space, achieving equivalent synchronization performance while reducing memory overhead and accelerating training convergence.

\noindent \textbf{Hybrid Mamba-Transformer Architecture.}
Previous methods \cite{difftalk_2023,Diff2Lip_2024,emo_2024} usually employ Transformer \cite{transformer_2017} to capture spatial and temporal patterns in video. 
Despite the impressive progress of these methods, the quadratic complexity of the attention mechanism in Transformer limits the image size and video length for inputs. Therefore, to further improve the computational efficiency of the model, this work replaces the self-attention layers in the denoising U-Net with Mamba-based layers \cite{mamba_2024, mambavision_2024} . 
For the input feature map $z\in\mathbb{R}^{b\times N \times c \times h \times w}$, the spatial Mamba handles the spatial dependencies along $h \times w$ dimension while the temporal Mamba operates on $N$ dimension. Attributable to the incorporation of Mamba layers, memory consumption for longer video inputs is reduced by 23\%, and the supported batch size is increased by 25\%, compared to a baseline model fully built with self-attention and cross-attention layers, which is similar to prior works MuseTalk \cite{musetalk_2024} and LatentSync \cite{latentsync2025}.
\section{Experiments}
\begin{figure*}[!t]
    \centering
    \includegraphics[width=1.0\textwidth]{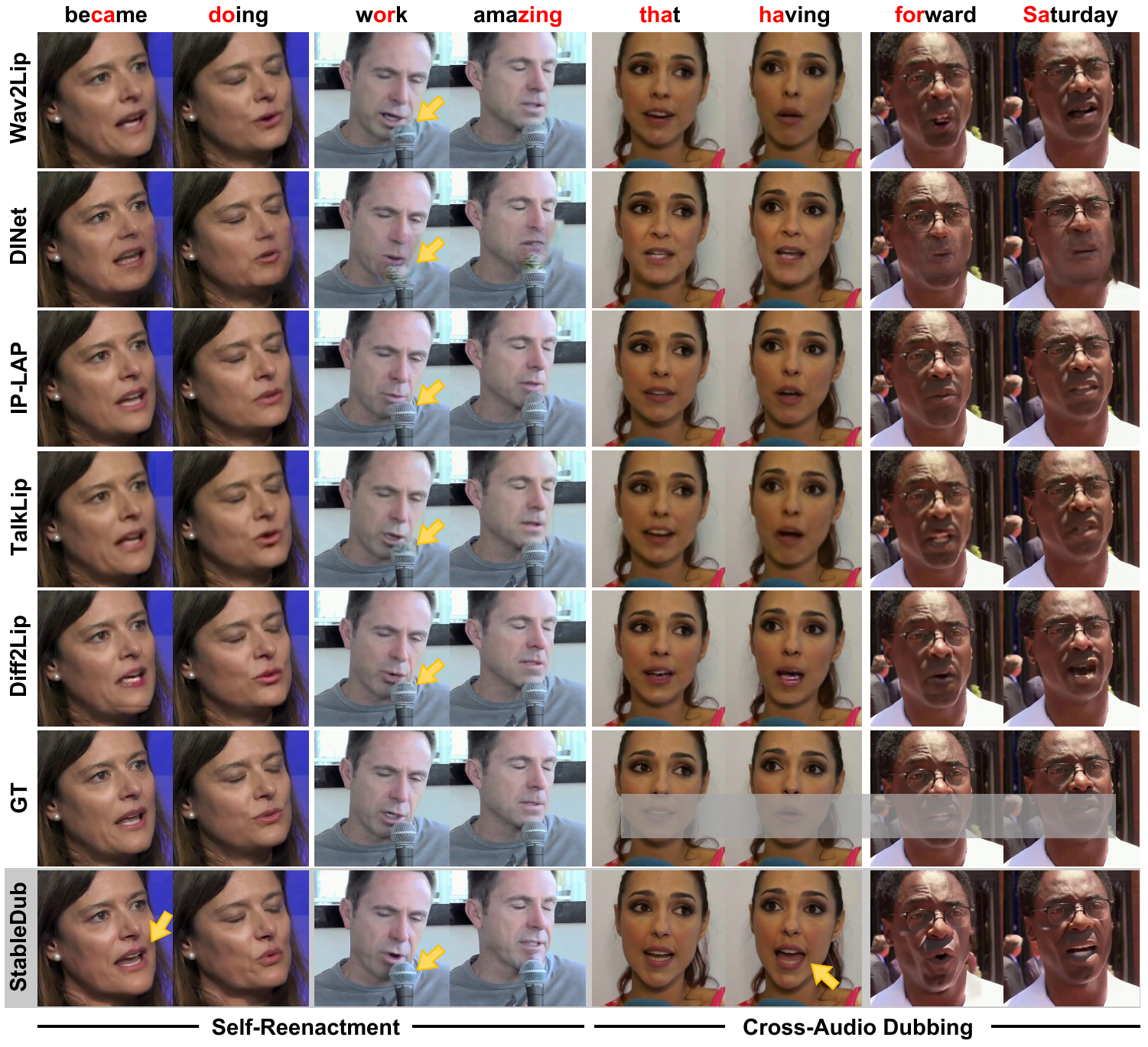}
    \caption{Qualitative comparison with other methods on the self-reenactment and cross-audio dubbing settings. There exists no ground truth for the cross-audio dubbing setting, so we mask the lower-half face. The first row displays the words that these frames are pronouncing. Please zoom in for clearer observation.}
    \label{fig:fig4}
\end{figure*}

\subsection{Experimental Settings}
\noindent \textbf{Training Dataset.} As shown in Fig.~\ref{fig:process}, we obtain a high-quality and diverse dataset by filtering VoxCeleb2 \cite{voxceleb2_2018}. All video clips are first resampled to 25 fps. Video clips of low clarity and with multiple people are directly discarded. 
We filter out clips with excessive head transitions, large head angles, or small facial sizes. Clips with low audio-lip synchronization and background noise are also removed. 
These preprocessing steps leave approximately 225 hours of paired audio-video data. We crop and resize each video clip to 512$\times$512.

\noindent \textbf{Test Datasets.} We evaluate methods on multiple datasets. Besides the commonly used HDTF \cite{hdtf_2021} and LRS2 \cite{lrs2_2017} datasets, we select more challenging video clips with high-resolution and difficult scenes (\textit{e.g.,} mouth occluded by objects) from the VFHQ dataset \cite{vfhq_2022} to test high-resolution video quality and generalization ability. We refer to this testset as VFHQ-C.
\begin{table*}[t]    
    \centering
    \small
    \renewcommand{\arraystretch}{1.1}  % 修改行间距
    \caption{Quantitative evaluation results on LRS2 dataset (WER metric) and VFHQ-C dataset (other metrics) under the settings of self-reenactment and cross-audio dubbing.}
    \resizebox{1\textwidth}{!}{
        \begin{tabular}{lccccccccccc}
            \toprule
            \multirow{2}{*}{Method} & \multicolumn{6}{c}{Self-Reenactment} & \multicolumn{5}{c}{Cross-Audio Dubbing}\\
            \cmidrule(lr){2-7} \cmidrule(lr){8-12}
            & LPIPS$\downarrow$ & LSE-C$\uparrow$ & LMD$\downarrow$ & FID$\downarrow$ & FVD$\downarrow$ & WER$\downarrow$ & LSE-C$\uparrow$ & LSE-D$\downarrow$ &FID$\downarrow$ &FVD$\downarrow$ &WER$\downarrow$\\
            \midrule
            Wav2Lip &0.1317 &7.2094 &0.813 &18.4620 &94.3733 &75.9 &6.4223 &7.8924 &18.5791 &110.8434 &76.3\\
            DINET &0.0955 &5.2746 &1.251 &12.3246 &115.8291 &78.4 &4.6266 &9.5506 &12.5959 &122.7919 &80.4\\
            IP-LAP &0.0795 &6.9510 &0.780 &10.7277 &63.8823 &66.5 &6.2773 &8.7868 &11.3303 &73.0300 &65.6\\
            TalkLip &0.1040 &6.7658 &1.117 &14.4779 &100.0698 &72.2 &4.8630 &9.5676 &14.8831 &111.7875 &75.8\\
            Diff2Lip &0.0931 &\textbf{7.5891} &0.851 &10.6108 &64.4021 &57.9 &6.3466 &7.5899 &11.3086 &82.5527 &68.6\\
            \hline
            Ground Truth &- &7.1041 &- &- &- &37.4 &6.9815 &7.4963 &- &- &36.2\\
            \rowcolor{gray!20} StableDub (Ours) &\textbf{0.0530} &7.1272 &\textbf{0.655} &\textbf{7.0433} &\textbf{34.8793} &\textbf{50.7} &\textbf{6.6421} &\textbf{7.4121} &\textbf{8.0915} &\textbf{45.0374} &\textbf{55.8}\\
            \bottomrule
        \end{tabular}
    }
    \label{tab:table1}
\label{tab:quantitative}
\end{table*}
\begin{figure}[t]
    \centering
    \includegraphics[width=0.46\textwidth]{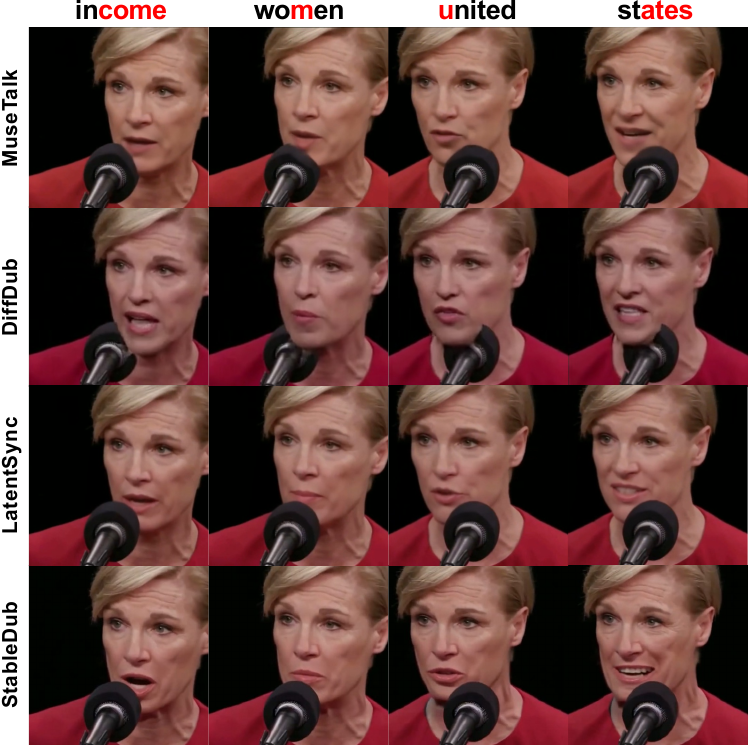}
    \caption{Visual comparison with the latest methods of similar architectures \cite{musetalk_2024,latentsync2025} and diffusion-based methods \cite{diffdub_2024, latentsync2025}.}
    \label{fig:diffusion_compare}
\end{figure}

\noindent \textbf{Implementation Details.} 
The training consists of two stages.
In the first stage, we reserve video with low sync since only images are needed. We train the spatial Mamba layer in each U-Net block with 36 million images to ensure the model generates high-quality and generalizable results.
In the second stage, we freeze the spatial Mamba and only optimize the audio cross-attention and temporal Mamba layers. The sequence length during training is set to $N=16$ to maximize the modeling of video consistency.
We adopt a similar U-Net design with Stable-Diffusion \cite{sd_2022}, except for the Mamba layer. To achieve faster convergence and leverage the strong generalization capability of Stable-Diffusion, we initialize the convolution and residual nets between the Mamba and attention layers and the cross-attention layer with the pre-trained weights of Stable-Diffusion. We train the model on 8 NVIDIA V100 GPUs. The training of the first stage takes 12 hours and the second stage takes 20 hours. Such training requirements are accessible for most researchers.

\noindent \textbf{Baseline Methods.}
The proposed method is for the person-agnostic visual dubbing task. As shown in Fig.~\ref{fig:teaser}, we compare StableDub with many code-available state-of-the-art (SOTA) methods. For conciseness, we report results of the five methods with the best overall performance. Their visualizations are presented in Fig.~\ref{fig:fig4}, while objective metrics and user study results are summarized in Tab.~\ref{tab:quantitative} and Tab.~\ref{tab:user_study}. 1) Wav2Lip \cite{wav2lip_2020} proposes to adopt a pre-trained lip-sync expert to enforce accurate, natural lip motion generation. 2) DINet \cite{dinet_2023}
performs spatial deformation on feature maps for better facial textural details. 3) IP-LAP \cite{iplap_2023} aligns multiple reference images with the target expression and pose for better appearance rendering. 4) TalkLip \cite{talklip_2023} adopts a lip-reading expert to improve the intelligibility of the generated lip regions. 5) Diff2Lip \cite{Diff2Lip_2024} is the latest audio-conditioned diffusion-based model for in-the-wild visual dubbing.

\begin{table}[t] 
    \centering
    \small
    % \fontsize{9pt}{11pt}\selectfont
    \renewcommand{\arraystretch}{1.1}  % 修改行间距
    \caption{User study measured by MOS with 95\% confidence intervals. VQ: visual quality. R\&S: lip habit resemblance and audio lip sync. OR: occlusion robustness. }
    \begin{tabular}{lccc}
        \toprule
        Method & VQ$\uparrow$ & R\&S$\uparrow$ & OR$\uparrow$ \\
        \midrule
        Wav2Lip &3.26$\pm$0.23 &3.59$\pm$0.29 &3.70$\pm$0.19\\
        DINET &2.48$\pm$0.24 &2.96$\pm$0.24 &2.62$\pm$0.22\\
        IP-LAP &3.89$\pm$0.23 &3.87$\pm$0.17 &3.61$\pm$0.20\\
        TalkLip &3.15$\pm$0.28 &3.20$\pm$0.29 &3.34$\pm$0.29\\
        Diff2Lip &3.58$\pm$0.21 &3.74$\pm$0.23 &3.77$\pm$0.21\\
        \hline
        Ground Truth &4.87$\pm$0.10 &4.82$\pm$0.15 &4.94$\pm$0.09\\
        \rowcolor{gray!20} StableDub &\textbf{4.71$\pm$0.18} &\textbf{4.67$\pm$0.20} &\textbf{4.81$\pm$0.14}\\
        \bottomrule
    \end{tabular}
    \label{tab:table2}
    % }
\label{tab:user_study}
\end{table}
\subsection{Quantitative Results}
We evaluate the self-reenactment and cross-audio dubbing tasks separately. For the former, the driving audio and input video are paired, while for the latter, the driving audio comes from another source.
The latter has no corresponding ground-truth video, which is closer to real applications.

\noindent \textbf{Evaluation Metrics.}
Following previous methods \cite{iplap_2023,Diff2Lip_2024}, we adopt the commonly used LPIPS \cite{lpips_2018} and FID \cite{fid_2017} to evaluate image quality. LSE-D and LSE-C by SyncNet \cite{syncnet_2017} and word error rate (WER) by an audio-visual speech recognition method \cite{avhubert_2022} measure the audio-lip sync. Landmark distance (LMD) \cite{lipsync3d_2021} on the mouth region can reflect the resemblance to the target lip habit and lip sync. To evaluate video consistency, we compute FVD \cite{fvd_2019} on 5-second video clips.

\noindent \textbf{Evaluation Results.}
The quantitative results are shown in Tab.~\ref{tab:table1}. The proposed method achieves the lowest LPIPS and FID scores for both self-reenactment and cross-audio dubbing tasks on the VFHQ-C testset, which demonstrates that StableDub obtains better generalization ability and generates images of higher quality. The proposed method also gains the lowest FVD score, which suggests that StableDub effectively preserves temporal consistency. For the audio-lip sync, we gain comparable and even better LSE-C and LSE-D scores than Wav2Lip, DINet, TalkLip and Diff2Lip, although they directly employ SyncNet for optimization. The lower WER score confirms the superior audio-lip sync of StableDub. It is worth noting that the proposed method shows more significant advantages in the cross-audio dubbing task, demonstrating its effectiveness in real applications.
The lowest LMD score in the self-reenactment task suggests that our method can effectively mimic the lip movement habit of the target avatar.
\begin{figure}[t]
    \centering
    \includegraphics[width=0.46\textwidth]{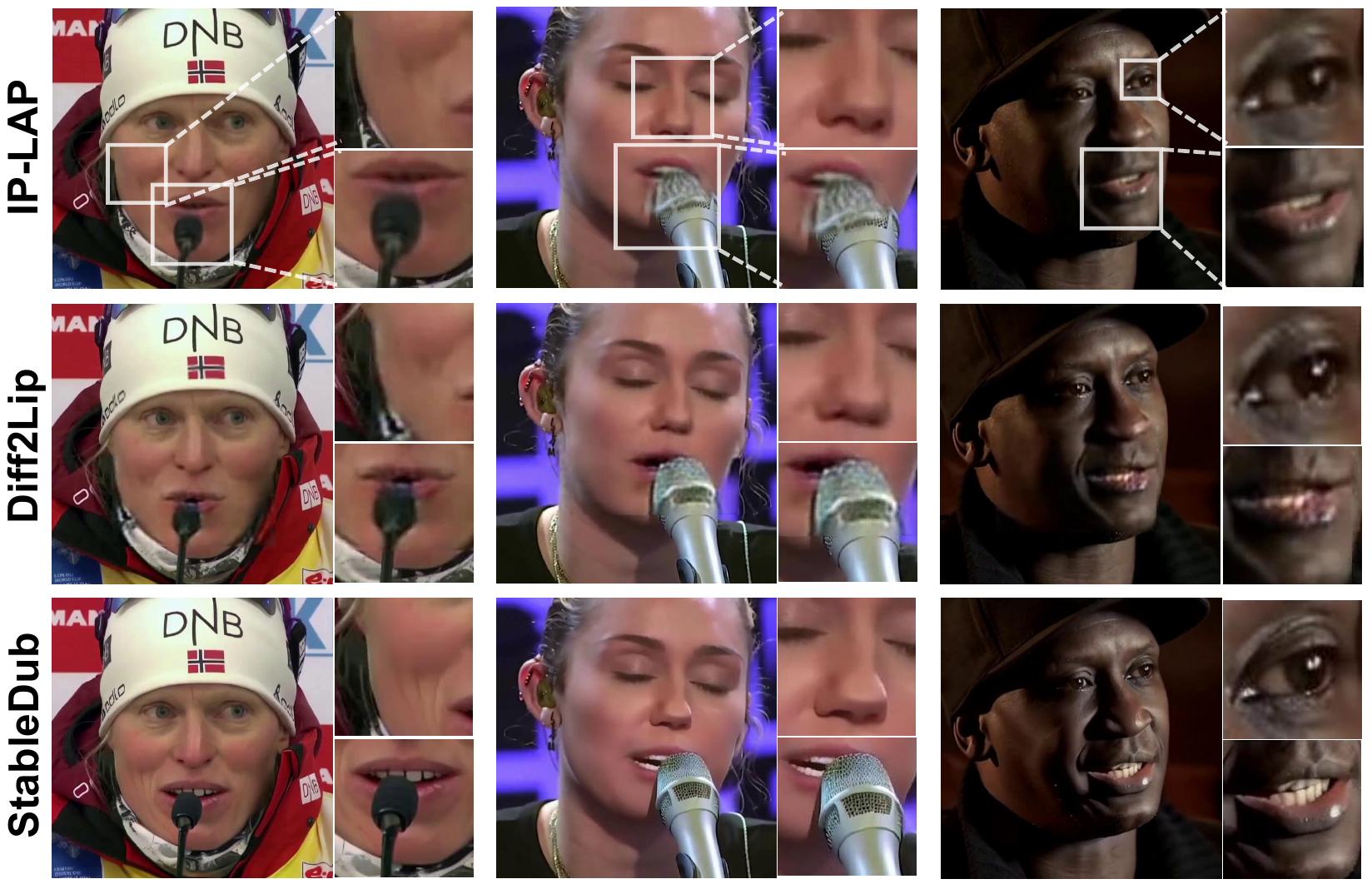}
    \caption{Visual comparison about object occlusion and uneven illumination under the cross-audio dubbing setting.}
    \label{fig:fig5}
\end{figure}

\noindent \textbf{User Study.} Current objective metrics for evaluating lip habit resemblance and occlusion robustness exhibit significant discrepancies compared to human perceptual assessment. We therefore employ user studies to better capture model performance. 
Thirty users with professional English ability participate in this study. The mean-opinion-score (MOS) tests are employed by asking users to score each video from one (worst) to five (best) on the aspects of visual quality, lip habit resemblance, audio-lip sync and occlusion robustness. As shown in Tab.~\ref{tab:table2}, IP-LAP and Diff2Lip achieve relatively high scores on visual quality and occlusion resemblance, while Wav2Lip can generalize to the challenging cases in VFHQ-C. The proposed method outperforms other methods on all aspects, proving the superiority of StableDub. The scores for lip habit resemblance and lip sync are remarkably close, thus we combine these two metrics. This also indicates that capturing lip habits contributes to enhancing lip sync.

\subsection{Qualitative Results}
\noindent \textbf{Visual Comparisons.} 
Fig.~\ref{fig:fig4} shows the visual comparison results. In terms of image quality, by closer inspection, our method is noticeably sharper and closer to the ground truth compared with other methods. Previous methods \cite{wav2lip_2020, Diff2Lip_2024} always have flaws in generating teeth, but StableDub can generate clear and realistic teeth. This results from the large-scale image pre-training in the first stage. For audio-lip sync and habit resemblance, our method shows more pronounced mouth opening and stronger expressiveness when pronouncing vowels in words like ``having'', ``doing'', and ``work'', proving that our method can generate accurate and vivid mouth movements. For the occlusion robustness, due to design flaws in masks and smaller generating areas, other methods often predict strange deformations of the microphone and mouth. However, our method generates a complete microphone and accurate mouth movements, handling the occlusion relationship between the mouth and microphone well. Fig.~\ref{fig:fig5} shows more details and also provides an example of the face under uneven lighting conditions to demonstrate stability.

To further demonstrate the superiority of our proposed method, we compare it with recent open-source methods, including MuseTalk \cite{musetalk_2024} (with a similar model architecture), DiffDub \cite{diffdub_2024} (diffusion-based), and LatentSync \cite{latentsync2025} (diffusion combined with a SyncNet expert), as illustrated in Fig.\ref{fig:diffusion_compare}. It can be observed that, with microphone occlusions, StableDub generates realistic and natural lip movements as well as complete occluding objects. Compared with LatentSync, StabldDub achieves superior audio-lip sync and visual quality with more expressive lip movements and clearer articulation of teeth.

\begin{figure}[t]
    \centering
    \includegraphics[width=0.45\textwidth]{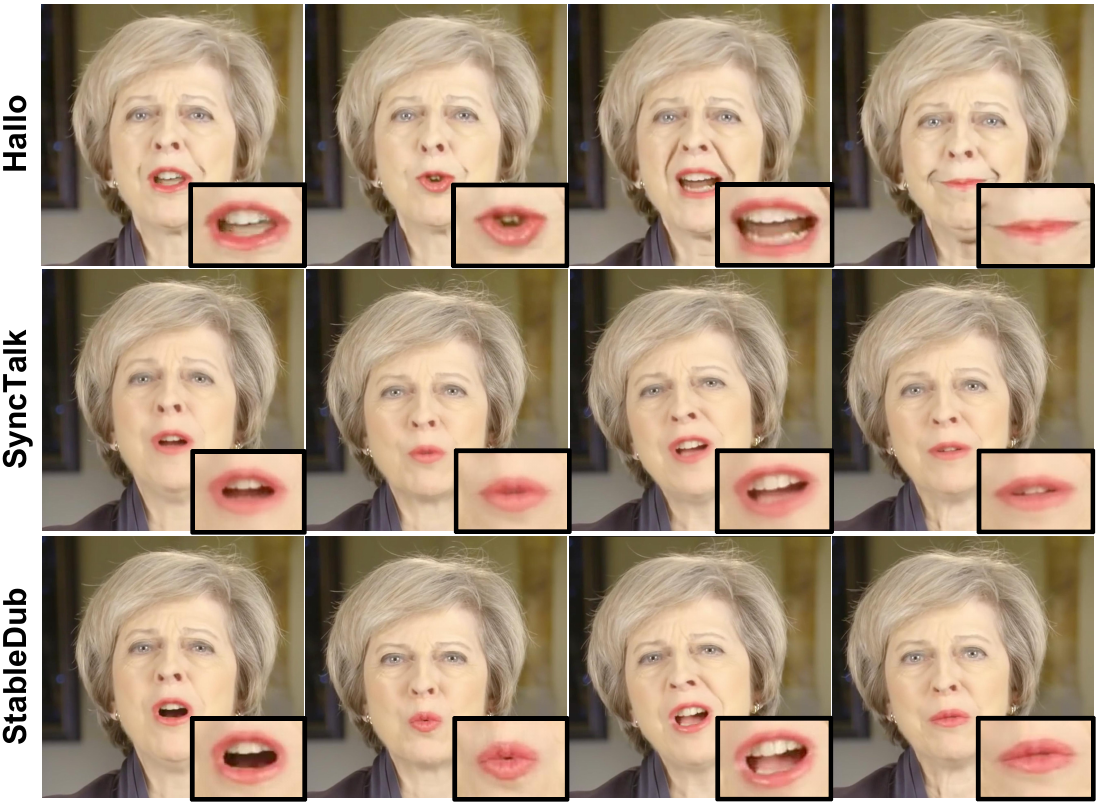}
    \caption{Comparison with dual U-Nets-based one-shot method Hallo \cite{hallo_2024} and person-specific method SyncTalk \cite{synctalk_2024}. SyncTalk shows the training results.}
    \label{fig:specific}
\end{figure}

As shown in Fig.~\ref{fig:specific}, we also compare with the SOTA person-specific method \cite{synctalk_2024} that captures the personalized lip habit during training, and the latest dual U-Nets-based one-shot animation method \cite{hallo_2024}. Our proposed method generates similar lip movements with SyncTalk, which verifies the habit modeling capability. Our method gains even better visual quality compared with Hallo. This shows the efficiency and generalization of StableDub.

\noindent \textbf{Visualization of Lip Habit Embeddings.}
% 我们从HDTF数据集中随机抽取了6个speaker，然后从他们的GT video提取了lip habit embedding。我们将这些embedding通过tsne的方式可视化。如图所示，不同speaker的habit embedding被分开，同一个speaker的embedding被聚类，这说明了我们的方法对于可以preserve target speaker specific的talking habit。
We randomly select 6 speakers from the HDTF dataset \cite{hdtf_2021} and extract corresponding lip habit embeddings from their ground-truth videos. As shown in Fig.~\ref{fig:tsne}, the t-SNE visualization \cite{tsne} of these embeddings reveals separation between different speakers' habit patterns while maintaining tight intra-speaker clustering. This empirical evidence demonstrates our method's capability to effectively preserve target-specific talking habits.

\begin{table}[t]
    \centering
    \small
    % \renewcommand{\arraystretch}{1.0}  % 修改行间距 1.15
    % \resizebox{0.5\textwidth}{!}{
    \caption{Ablation study on generalization ability. The \underline{operations} that cause the most significant performance degradation in each aspect are underlined.}
    \begin{tabular}{lccc}
        \toprule
        Setting & FID$\downarrow$ & FVD$\downarrow$ & LMD$\downarrow$ \\
        \midrule
        StableDub &\textbf{7.0433} &\textbf{34.8793} &\textbf{0.655}\\
        \midrule
        Dual U-Nets &7.0519 &34.8904 &0.692\\
        w/o Mask Exposure  &7.1046 &36.5170 &{0.725}\\
        w/o Habit Modulation &7.0579 &34.9100 &\underline{0.810}\\
        Mamba $\rightarrow$ Attention &\underline{7.1704} &36.5026 &0.691\\
        w/o Temporal Mamba &7.1109 &\underline{49.2208} &0.703\\
        \bottomrule
    \end{tabular}
    \label{tab:table3}
    % }
\end{table}
\begin{figure}[t]
    \centering
    \footnotesize
    \includegraphics[width=0.4\textwidth]{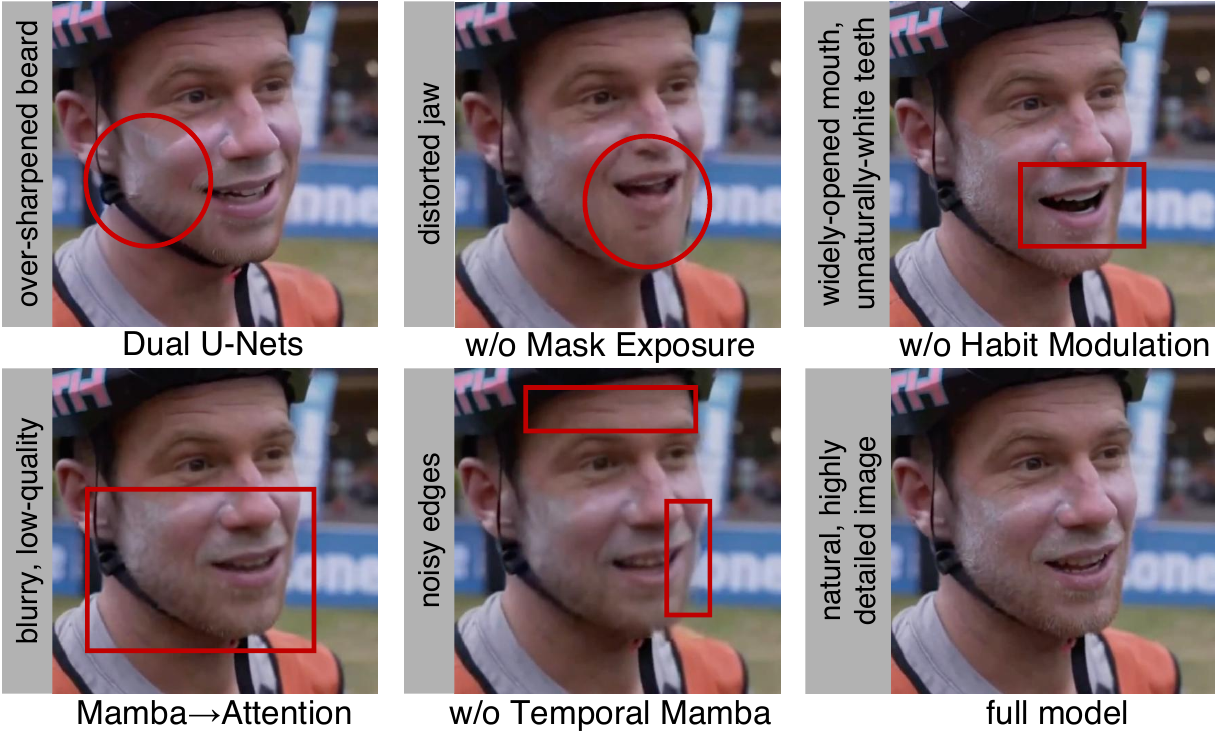}
    \caption{Visualization for ablation study. Our full model generates high-quality and detailed images.}
    \label{fig: ablation_vis}
\end{figure}
\begin{figure}[t]
    \centering
    \includegraphics[width=0.49\textwidth]{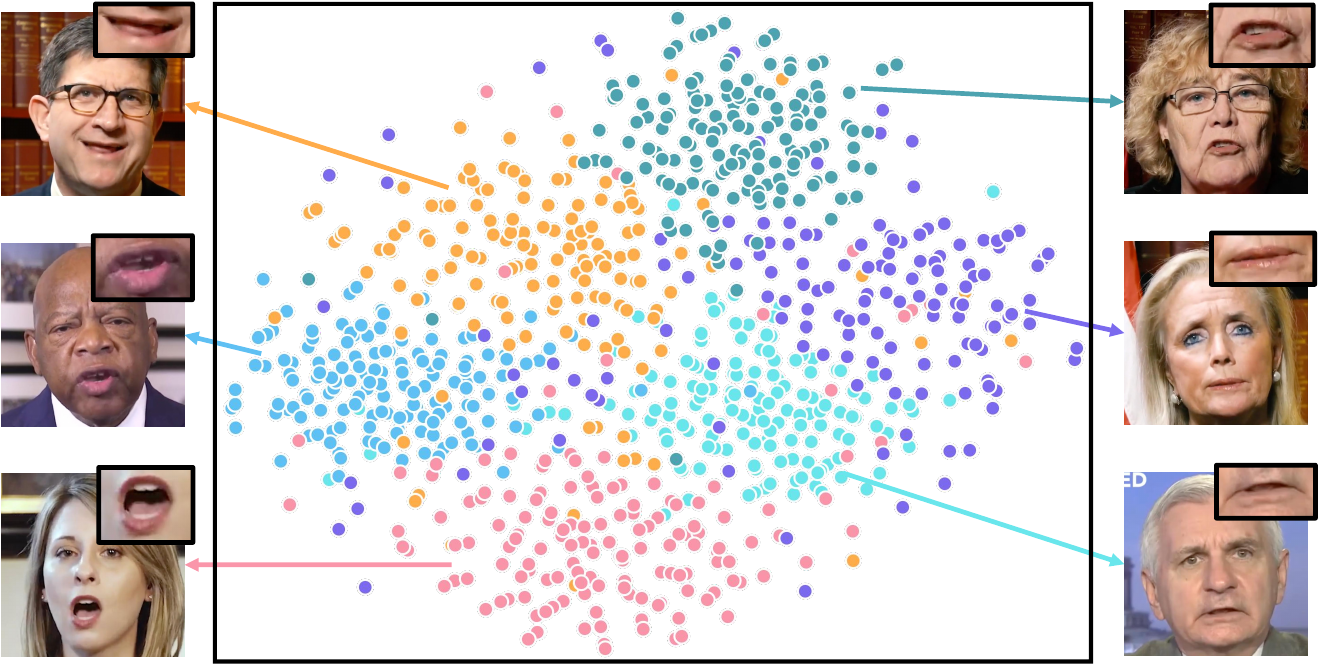}
    \caption{The t-SNE visualization of extrated lip habit embeddings. Points of different colors indicate different speakers.}
    \label{fig:tsne}
\end{figure}

\begin{figure}[t]
    \centering
    \includegraphics[width=0.46\textwidth]{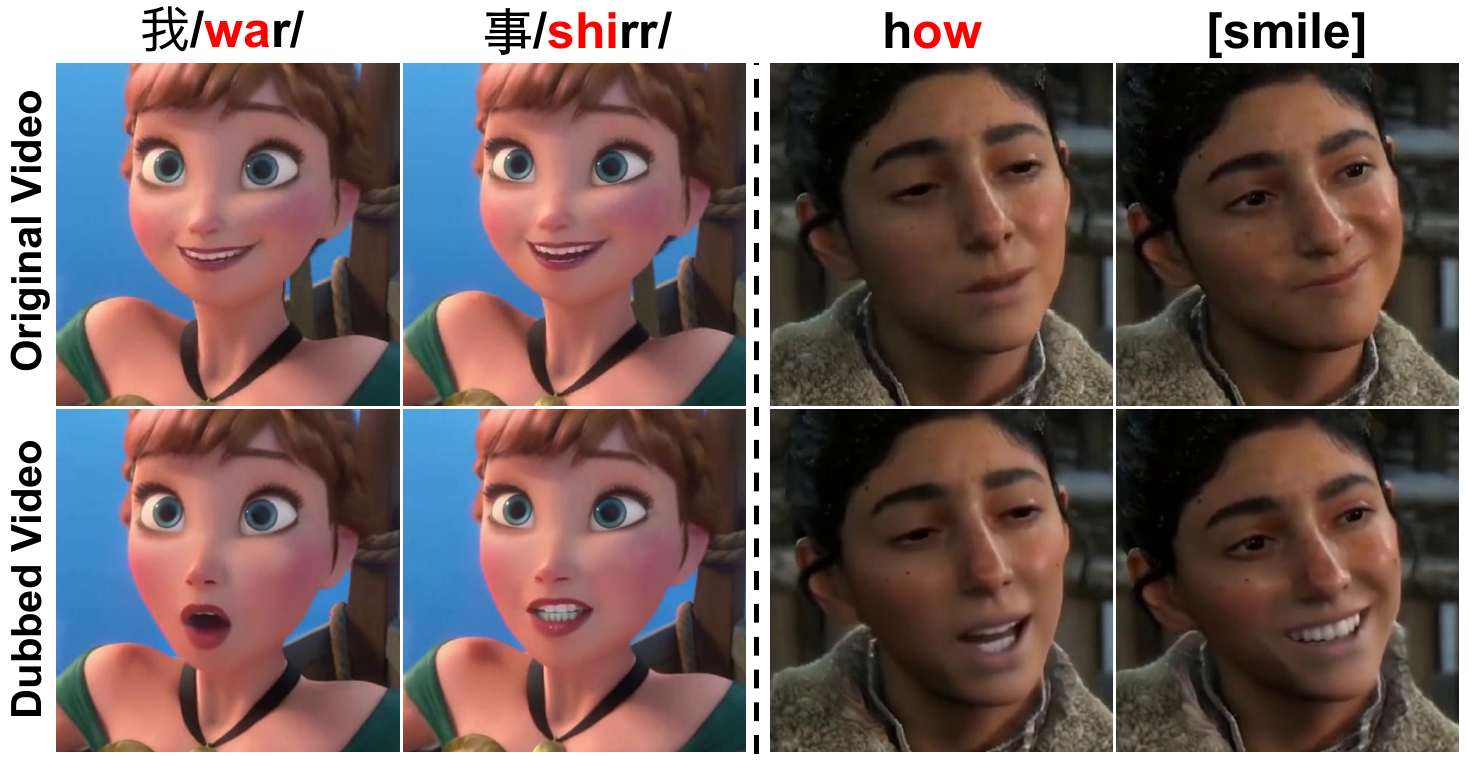}
    \caption{Out-domain generation. Left: Chinese dubbed ``\emph{Frozen}''. Right: original audio redubbed ``\emph{The Last of Us}''. StableDub corrects motion capture artifacts in the original video.}
    \label{fig:fig6}
\end{figure}

\noindent \textbf{Out-Domain Generation.}
StableDub also demonstrates strong stability for out-of-domain audio and facial images. As shown in Fig.~\ref{fig:fig6}, we test our method on ``\emph{Frozen}'', an American English animated musical fantasy film. We input the Chinese singing audio for testing. Our method not only inpaints the images with the film style but also matches the lip movements with the Chinese singing. This indicates that our method can perform video translation or dubbing beyond realistic human portraits. 
We also test our method in an interesting scenario of redubbing the characters in the action-adventure 3D game ``\emph{The Last of Us}'', since traditional game creation usually employs motion capture for lip-driving, which is expensive and lacks accuracy. Our method not only generates more accurate and expressive lip movements than the original video but also provides realistic and natural detail enhancements in regions like teeth. This result shows that our method has the potential to be used in pipelines for refining 3D game character animations to enhance user experience.

\subsection{Ablation Study}
% To further validate the effectiveness of the contributions proposed in this paper, we conduct an ablation study and compute the relevant quantitative metrics.

\noindent \textbf{Generlization Ability.} 
As shown in Tab.~\ref{tab:table3}, we test different model settings separately. 1) We adopt the dual U-Nets structure \cite{animateanyone_2024, emo_2024, hallo_2024} with an additional referenceNet to extra and inject appearance features. It can be observed increasing the parameter count does not result in a significant performance improvement.
2) We remove the strategy of mask exposure. The rise in LMD indicates that the strategy also results in better lip sync. 3) Removing the lip habit modulation module leads to a significant LMD rise. It proves the necessity for lip habit awareness. 4) Replacing all Mamba layers with attention results in degraded image quality. We attribute the quality gains to Mamba’s ability to support larger batch sizes and longer sequences during training. More stable optimization and extended video context contribute to improved visual fidelity. 5) Removing the temporal Mamba layer increases FVD, leading to a decline in video consistency.
Fig.~\ref{fig: ablation_vis} shows the visual results of the ablation study. Removing the strategy of mask exposure leads to distorted images.
\begin{figure}[t]
    \centering
    \includegraphics[width=0.43\textwidth]{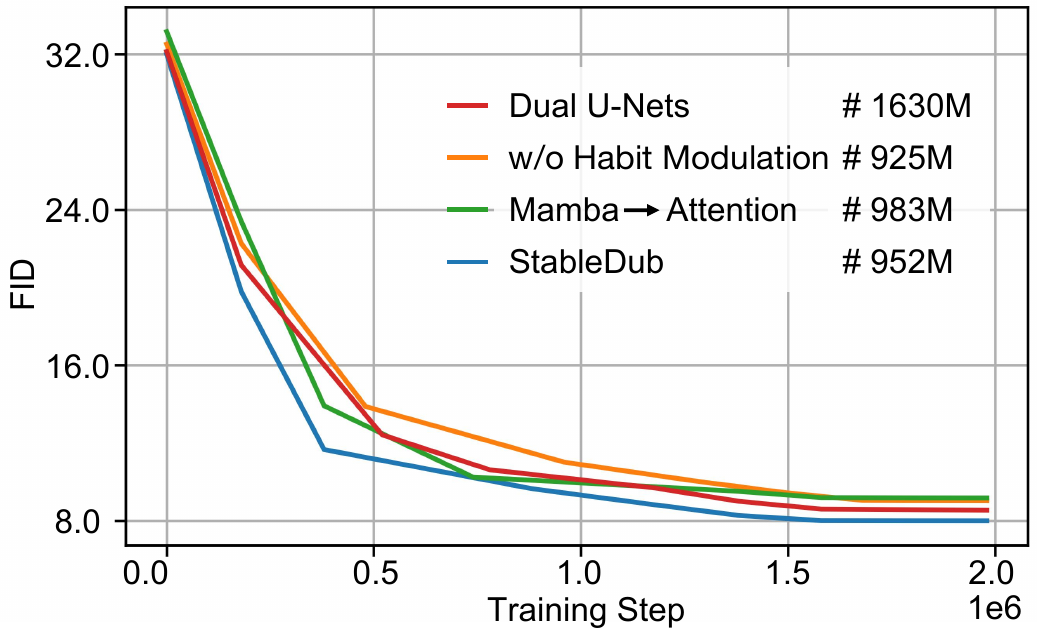}
    \caption{Curves of training convergence. The parameter counts are given next to the corresponding model structures.}
    \label{fig:train_curve}
\end{figure}
\begin{figure}[t]
    \centering
    \includegraphics[width=0.48\textwidth]{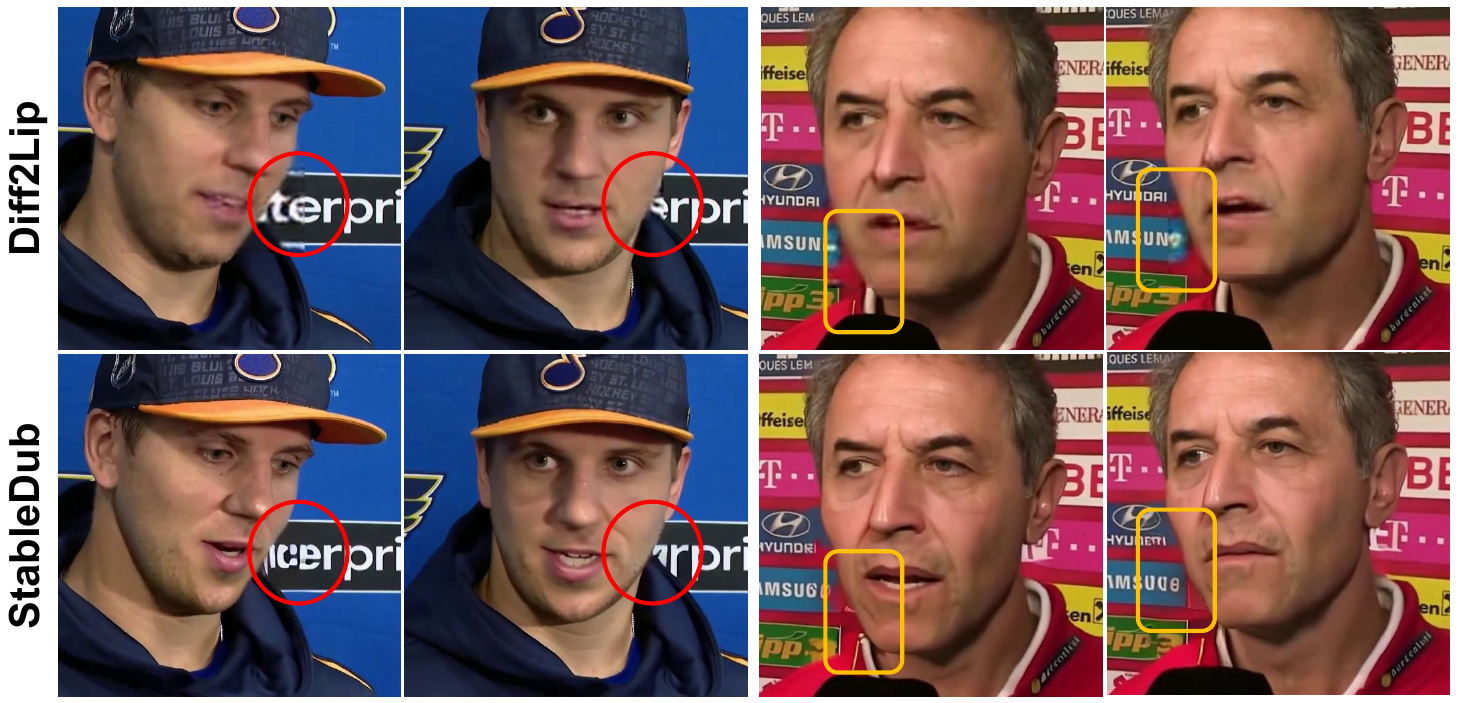}
    % \vspace{-8pt}
    \caption{Failure cases of StableDub and Diff2Lip.}
    \label{fig:supp3}
\end{figure}

\noindent \textbf{Training Efficiency.}
Fig.~\ref{fig:train_curve} shows the curves of training convergence for different model structures and their corresponding parameter counts.
The dual U-Nets structure has the largest number of parameters, but under the same training steps, it struggles to converge to the same FID value as StableDub. Replacing Mamba with attention results in a significant decline in convergence speed, demonstrating the advantages of the hybrid Mamba-Attention net. Removing the lip habit modulation module also leads to the slower convergence speed of FID. 
This is due to the absence of a reference for the generation of the lip region, requiring a longer optimization time. 
As shown in Table \ref{tab:diff}, our method demonstrates superior efficiency by eliminating redundant priors in other works, employing a more concise framework, and reducing training overhead, while simultaneously achieving enhanced generation performance.
In line with the discussion in Sec. \ref{sec:Mamba-Transformer}, the proposed method leverages a relatively large network parameter count and effectively balances generalization and training efficiency.

\section{Limitations}
\label{sec:limitation}
% StableDub在person-agnostic任务上取得了state-of-the-art 结果，并且在unseen的和out-of-domain的数据上取得了出色的泛化效果，但它在以下几个方面仍存在缺陷和提升空间。1）从图中可以看出，任务的背景没有被很好的保留，这是因为我们的方法使用了更大的mask，生成了更大区域，这一方面保证了cross-frame的时序一致性，另一方面也使得方法对复杂背景的泛化性下降。我们将在未来探索两种解决方法，一是appearance net的输入即是当前input video的image frame，通过appearance net将复杂背景信息传给denoising net； 二是我们将联合使用不同尺度的mask使网络在temporal consistency和background preservation之间自动做权衡。2）尽管diffusion 方法能够取得比GAN更好的生成质量，但是在推理速度方面它却远远落后于GAN。我们相信对diffusion加速方法对探讨和引入将进一步扩大我们方法的应用范围
StableDub achieves SOTA results in the person-agnostic visual dubbing task, but it still has some limitations. (1) As shown in Fig.~\ref{fig:supp3}, the background is not well preserved. This is because our method employs a larger mask and renders a larger region, which reduces the perception of complex backgrounds. Another contributing factor is that only a single reference image is used. Employing multiple reference images could mitigate this issue. We will explore jointly adopting masks of different scales and leverage multiple reference images for improved visual fidelity and generalizability. (2) Although diffusion methods can achieve better generation quality than GANs \cite{wav2lip_2020,iplap_2023}, the inference speed is inferior to that of GANs. We believe that exploring acceleration methods \cite{lcm_2023} for diffusion networks will further expand the application scope of our method.
\section{Conclusion}
This paper focuses on enhancing the applicability of visual dubbing methods through improvements in generalization and efficiency. The proposed lip habit-aware modeling enables the generated avatar to preserve the original speaker's articulatory patterns, while occlusion-aware inpainting ensures natural visual outputs even under occluded scenarios. These generalization enhancements eliminate the reliance on redundant prior required by previous methods. Combined with the hybrid Mamba-Transformer architecture, the framework achieves enhanced computational efficiency. Extensive experiments and analyses demonstrate that StableDub outperforms other methods across multiple dimensions. This comprehensive advancement improves the practical applicability of visual dubbing systems for real-world deployment.

\normalem
\bibliography{main}
\bibliographystyle{IEEEtran}

\vfill

\end{document}